\documentclass[utf8]{FrontiersinVancouver} 

\setcitestyle{square} 
\usepackage{url,hyperref,lineno,microtype,subcaption}
\usepackage[onehalfspacing]{setspace}

\def\keyFont{\fontsize{8}{11}\helveticabold }
\def\firstAuthorLast{Delahunt {et~al.}} 
\def\Authors{Charles B. Delahunt\,$^{1,*}$, Noni Gachuhi\,$^{1}$ and Matthew P. Horning\,$^{1}$}
\def\Address{$^{1}$Global Health Labs, Bellevue, WA, 98007, USA} 
\def\corrAuthor{Charles Delahunt}

\def\corrEmail{charles.delahunt@ghlabs.org}

\begin{document}
\onecolumn
\firstpage{1}

\title {Metrics to guide development of machine learning algorithms for malaria diagnosis} 

\author[\firstAuthorLast ]{\Authors} 
\address{} 
\correspondance{} 

\extraAuth{}

\maketitle

\begin{abstract}

Automated malaria diagnosis is a difficult but high-value target for machine learning (ML), and effective algorithms could save many thousands of children's lives. 
However, current ML efforts largely neglect crucial use case constraints and are thus not clinically useful. 
Two factors in particular are crucial to developing algorithms translatable to clinical field settings: (i) Clear understanding of the clinical needs that ML solutions must accommodate; and (ii) task-relevant metrics for guiding and evaluating ML models.
Neglect of these factors has seriously hampered past ML work on malaria, because the resulting algorithms do not align with clinical needs.

In this paper we address these two issues in the context of automated malaria diagnosis via microscopy on Giemsa-stained blood films. 
First, we describe why domain expertise is crucial to effectively apply ML to malaria, and list technical documents and other resources that provide this domain knowledge.
Second, we detail performance metrics tailored to the clinical requirements of malaria diagnosis, to guide development of ML models and evaluate model performance through the lens of clinical needs (versus a generic ML lens). 
We highlight the importance of a patient-level perspective, interpatient variability, false positive rates, limit of detection, and different types of error. 
We also discuss reasons why ROC curves, AUC, and F1, as commonly used in ML work, are poorly suited to this context. 
These findings also apply to other diseases involving parasite loads, including neglected tropical diseases (NTDs) such as schistosomiasis.  
  
\tiny
 \keyFont{ \section{Keywords:} Malaria, NTDs, schistosomiasis, metrics, machine learning, sensitivity, specificity, limit of detection, ROC, AUROC, hemozoin} 
\end{abstract}

\section{Introduction}
\label{sec:introduction}

Malaria and some neglected tropical diseases (e.g., schistosomiasis) involve parasite loads that can be detected in microscopy images of a substrate (e.g., blood or filtered urine). They are thus amenable, though difficult, targets for automated diagnosis via machine learning (ML) methods. 
These diseases are also very high-value ML targets: They are serious global health challenges affecting hundreds of millions of people, especially children, in underserved populations \citep{whoMalaria2019, gfNTDs}. 

\let\thefootnote\relax\footnotetext{\copyright $~~$ Global Health Labs, Inc.}

\begin{figure}[h!]
\begin{center}
  {\includegraphics[width=0.8\textwidth]{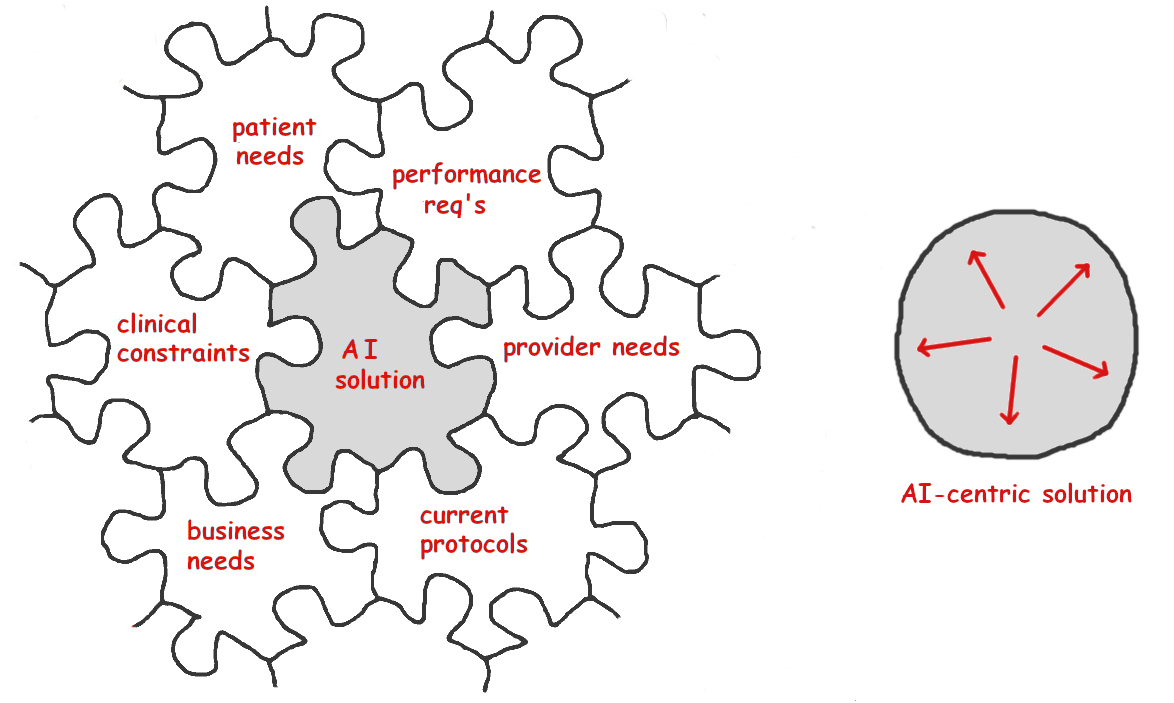}}
  \end{center}
  {\caption{Left: Effective ML (AI) solutions must interlock with domain requirements and will be shaped by non-ML pressures from the use case. Right: Solutions developed with a ML-centric approach, neglecting the use case, will fail to match clinical needs. (``interlocking" metaphor due to Dr. Scott McClelland; jigsaw outline from \citep{draredech})}
  \label{fig:jigsaw}
  }
\end{figure}

However, ML methods developed for malaria diagnosis using Giemsa-stained blood films have so far largely failed to translate to useful deployment, for several reasons. \\
$~~~~$(i) The task is difficult: for example, malaria parasites are small; field blood films are highly variable and often full of distractor objects; and the low limits of detection required for clinical use result in low signal-to-noise ratios (e.g., one parasite per 30 large fields of view). \\
$~~~~$(ii) ML development has typically proceeded in a heavily ML-centric mindset, without careful attention to (or even knowledge of) the domain specifics, use cases, and clinical requirements of malaria. 
This yields algorithms that, almost by design, fail to meet clinical needs and cannot be built upon (see Figure \ref{fig:jigsaw}). \\
$~~~~$(iii) ML development can only optimize what is measured, so a crucial prerequisite for successful development is a set of task-relevant metrics \citep{metricsReloaded}. 
These tailored metrics have largely been lacking for malaria, for which ML development has instead been guided by generic and ill-suited ML metrics such as object-level ROC curves.
This paper seeks to accelerate the ML community's progress towards  translatable solutions for malaria diagnosis, by describing tools and techniques which we have found to be essential for development of clinically effective ML algorithms.
It captures lessons learned by our group over a decade of applying ML to malaria diagnosis. 
The resulting algorithms \citep{ghtc2014, mehanian, thinGhtc2019} represent, to our knowledge, the most effective and also most extensively field tested \citep{torres, vongpromek, horningWho55, das, reesChanner} ML algorithms yet built for fully automated diagnosis of malaria on Giemsa-stained blood films.
These field trials showed that our  algorithms, though state-of-the-art, still fall short of the clinical demands, and highlight the need for more robust algorithms to truly impact malaria diagnosis.

The paper is structured as follows: Section 2 details aspects of ML work that depend on a grasp of the clinical use-case (e.g., how the disease is diagnosed in the field), lists malaria documents especially relevant to ML work, and discusses other domain knowledge resources.
Section 3 describes ML metrics tailored specifically to malaria and NTDs that can be applied during development of ML algorithms to optimize their clinical effectiveness, and also describes problems with some commonly-used ML metrics.
%
%
\section{The clinical use case}
\label{sec:theClinicalUseCase}
To be clinically useful an ML solution must fit into a larger, ML-independent context.
It must interlock with other pieces that are shaped by clinician needs, site requirements, protocols currently in use, patient needs, business environment, etc. \citep{wiens}.
This strong constraint to mesh with non-ML considerations is often overlooked by ML practitioners, leading to algorithms that are elegant (from an ML perspective) but useless (from a clinical perspective) \citep{koller}.

In particular, a clinically useful ML algorithm must fit into an existing care structure and meet or exceed existing clinical performance targets. 
So understanding these clinical constraints is a basic prerequisite for algorithm development.  
(We set aside the complex case of a disruptive technology potentially altering existing care protocols. Such cases of course require careful analysis.) \\
$~~~$This section discusses some crucial points to consider, and lists resources for learning about malaria use cases.
\subsection{Important domain specifics}
\label{subsec:importantDomainSpecifics}
Several domain-specific details are fundamental to effective algorithm development:\\ \\
\textit{Basic facts about the clinical needs:}\\
For example, what are the proper uses of thick vs. thin blood films for malaria?  \\ \\
\textit{Performance metrics relevant in the clinic:} \\
Examples include patient-level sensitivity and specificity, and limit of detection (LoD).
This knowledge enables ML researchers to tailor salient metrics to guide algorithm development (like those we give in Section \ref{sec:salientMetrics}), define objective functions, do internal assessment, and report algorithm results meaningfully. \\ \\
\textit{Performance specifications:}\\
Clinicians are unwilling to reduce patient care standards, so ML models must perform at least as well as current practice to be deployable.
Field performance requirements are thus vital concerns, even if a particular model iteration does not attain them (since the work can then be built upon or extended). \\ \\
\textit{Domain-specific obstacles and shortcuts:} \\
Some difficult details need special treatment, and others allow for valuable shortcuts.
For example, malaria parasites can exist at various depths of a thick blood film, so a single image plane will not capture all parasites in focus. 
On the plus side, the nuclei of white blood cells (WBCs) are plentiful in thick films and stain similarly to malaria parasite nuclei, so they can serve as a ready-made color reference for the rare (or absent) parasites.
Shortcuts matter because generic methods applied as-is are unlikely to hit clinical performance requirements, which is a much harder task than simply outdoing another generic method in a ML-style comparison. \\ \\
\textit{Structuring annotations and training sets:}\\
Annotations and training data are central to ML success, and must be tailored to the task.
For example, malaria ring forms (the youngest parasite stage) typically have both a round nucleus and a crescent-shaped cytoplasm (examples in Fig \ref{fig:intersampleVariability}). 
However, after drug treatment the rings often lack visible cytoplasm, appearing in thick films as dark round dots which are very similar to a common distractor type. 
As a result, they have outsized impact on decision boundaries and require special care as to annotation and inclusion in training sets. 

Avenues to acquire vital domain expertise include (i) documentation and (ii) connecting with domain experts.
 
\subsection{Documentation} 
 
Effective ML solutions need to design in accommodations to non-ML (e.g., clinical) constraints.
Therefore, literature review to inform ML work should extend well beyond ML methods and focus on the clinical use-case itself, without an ML-centric filter.
Documentation of use cases and standards of care are published by various agencies, including the World Health Organization (WHO), ministries of health, and non-government organizations (e.g., the Bill and Melinda Gates Foundation, 
the Global Fund, 
and the Worldwide Antimalarial Resistance Network).

Below we list some references that are especially relevant to ML researchers designing algorithms for automated malaria diagnosis using Giemsa-stained blood films. \\ \\ 
\textit{Appropriate evidence for ML:} \\
$~~~${\tiny\textbullet} The WHO has issued guidelines on how to generate meaningful evidence for ML-based medical tools \citep{whoEvidenceForAi}, especially Section 1.
This document is important for ML as applied to any medical use case.
Crucially, evidence of algorithm performance during development must be firmly grounded in the clinical use case. This requirement underpins the metrics described below in \ref{subsec:patientLevelMetrics} - \ref{subsec:speciesIdMetrics}. \\ \\
\textit{Protocols for malaria microscopy:} \\
Various groups have published diagnosis protocols which detail the clinical task.\\
$~~~~${\tiny\textbullet} WHO's guidelines are an essential resource \citep{whoMicroscopyLearners2010} and \citep{whoMicroscopy2016} (see especially SOPs 8 and 9 for diagnosis and quantitation).  \\
$~~~~${\tiny\textbullet} Ministries of Health also have useful protocols, e.g., Peru \citep{peruMalaria} and USA \citep{cdcMalaria}.\\
$~~~~${\tiny\textbullet} WWARN and the WHO have developed protocols tailored to research contexts (e.g., drug resistance sentinel sites) \citep{whoMalariaResearch2016}.\\ \\
\textit{Evaluation tests:} \\
$~~~~${\tiny\textbullet} The WHO has developed a system to evaluate malaria microscopists.
This uses a set of 56 blood slides with carefully specified parasitemias and species \citep{whoQualityAssuranceV2} (section 6).
The ``WHO 56" evaluation reflects the tasks and accuracies required in the clinic and is thus a valuable and challenging test for ML algorithms.
Its difficulty gives an appreciation of the skills of human field microscopists.
The defined competency levels offer clear and clinically meaningful performance targets for ML algorithms.  
Note that the ``WHO 56" differs slightly from the previous version (the ``WHO 55") found in \citep{whoQualityAssuranceV1}.\\
$~~~~${\tiny\textbullet} A similar but distinct evaluation set of blood slides, tailored to research rather than clinical contexts, is detailed in \citep{whoMalariaResearch2016}.\\
$~~~~${\tiny\textbullet} Peru's quality control protocols implicitly describe performance requirements \citep{peruMalaria}, sections 7.2 and 9.2.\\ \\
\textit{Neglected tropical diseases:} \\
$~~~~${\tiny\textbullet}  The WHO has defined target product profiles, including sensitivity and specificity requirements, that are relevant to automated ML systems targeting schistosomiasis \citep{whoHelminths}. \\ \\
\textit{Other performance specifications:} \\
$~~~${\tiny\textbullet} The above documents also provide detail concerning other general product requirements relevant to any ML solution that aims for translation to clinics. 
These issues include time-to-result, throughput, electricity/battery constraints, price, and (implicitly) computational constraints. \\ \\ 
\textit{ML publications:}\\
$~~~~${\tiny\textbullet} Some ML papers (e.g., \cite{horningWho55,oyiboSchisto2022}) cite non-ML documents relevant to use case, but this is not (yet) common practice.
So ML-based literature search is insufficient.
%
\subsection{Domain experts}
\textit{Domain experts} are a vital source of guidance and collaboration. 
They include field experts, i.e. those who work in field clinics or who do field-based research; and subject matter experts, such as WHO personnel and long-time researchers in the space (these groups overlap).
The value of their experience and insight to effective algorithm development cannot be overstated.

As an example, our group's entire ML program for malaria diagnosis has depended absolutely upon expert input from a technical advisory panel, as well as on continued contacts and advice from field clinics.
To the degree that our work has succeeded, this expert input has been the key ingredient (along with the closely entwined matter of data collection and curation).
We would argue that ML development can only progress towards clinically useful algorithms when domain expertise is somehow integrated into the team (recent examples include \citep{manescu2020, manescuWeaklySupervised, kassim, poostchi2022, yang2020, yu2023} and for schistosomiasis  \citep{oyiboSchisto2022, armstrong})

Connecting with such experts is made easier by two things. 
First, people (on average) love to talk about their work.
Second, field experts are often (again, on average) open to engaging with ML solutions and happy to co-author serious research.

Sources for contacts include: (i) published work, e.g., who is leading and authoring/co-authoring relevant studies; (ii) academic institutions with concentrations of research in the space; (iii) online interest groups, e.g., on LinkedIn; and (iv) non-ML conferences, their attendees, and proceedings, e.g., the American Society of Tropical Medicine and Hygiene. 


\section{Salient metrics}
\label{sec:salientMetrics}

Salient metrics are essential to ML work, both to guide development and to report results meaningfully. 
Unfortunately, the metrics routinely applied to ML work on malaria (e.g., object-level precision, recall, AUC, and F1 score) have disqualifying drawbacks in the malaria context.

A 2018 review of automated malaria detection papers \citep{poostchi} described serious problems (which still persist): 
reported metrics are incomplete and not comparable between studies; 
metrics are object-based (not patient-based) and are thus not relevant to the clinical task; 
train and test sets contain objects from the same patient, which  contradicts the patient-level focus;
and datasets are too small.
In addition, incorrect assumptions are built into algorithms: for example, diagnosis on thin blood films is common in ML papers, despite being contrary to clinical practice due to practical obstacles \citep{whoMicroscopy2016, earlLongComment2015} (though see recent work on thin film spreaders in \citep{noul, prakeshThinFilmSpreader}).

In this section, we first discuss some of the problems with commonly-used ML metrics (\ref{subsec:problemsWithRocsAndAucs}). 
We then describe in detail some alternate metrics which have high clinical relevance for the malaria use case (\ref{subsec:patientLevelMetrics} - \ref{subsec:speciesIdMetrics}).
These metrics are effective tools both to guide ML development and to report meaningful performance results.
They are suitable for ML models that target diseases involving parasite loads such as malaria, NTDs, or more generally any pathology where diagnosis is determined by the presence of a variable number of abnormal objects (e.g., pixels or cells in a histopathology slide).
 

\subsection{Problems with ROCs, AUCs, and Precision}
\label{subsec:problemsWithRocsAndAucs}
 
ML practitioners choose metrics to evaluate model performance by (i) what is customary, familiar, and convenient; (ii) what has been done by previous authors; (iii) what can generate the ``state of the art" (SOTA) comparisons required for publication in the ML community; and (iv) what is acceptable to ML reviewers.  
This creates a closed loop which perpetuates the use of certain metrics without regard to their effectiveness. 
When entrenched metrics do not assess algorithm performance in a clinically relevant way, it blocks progress towards deployable solutions.

Several commonly-used ML metrics, including object-level ROC curves, AUC, object precision, and F1 score, appear frequently in the ML malaria literature. 
But in the malaria context these are flawed measures of performance, inappropriate as evidence per \citep{whoEvidenceForAi}, and should therefore be avoided (they can be useful intermediate measures for internal algorithm work).
\subsubsection{Object-level ROC curves and AUC}
\textit{Object-level ROC curves}, and the associated Area Under Curve (AUC), are routinely reported by ML research papers involving parasite detection.
However, they have three key weaknesses in this context (except perhaps as intermediate measures for internal algorithm work).

First, they do not address the clinical need for patient-centric care.
In particular, they ignore the crucial matter of patient-level variability of object-level accuracy (this variability is discussed in \ref{subsec:sensitivity} and \ref{subsec:fpRate}). 

Second, real samples often have a large imbalance between distractors and positive objects, especially at parasitemias near clinical LoD.
A common situation is a model that diagnoses malaria on thin films by labeling individual RBCs as infected or not. 
5 million RBCs/$\mu L$ vs. $100$ p/$\mu L$ at LoD gives 50,000 negative objects for each positive object, so a 0.999 AUC can coexist with an average of 50 FPs \textit{per parasite} at LoD (a very low SNR).
Since one detected parasite and one FP object have equal impact on diagnosis (as determined by exceeding a threshold $T$), FP noise will swamp the diagnostic signal of detected parasites.

\begin{figure}[h!]
\begin{center}
{\includegraphics[width=0.9\textwidth]{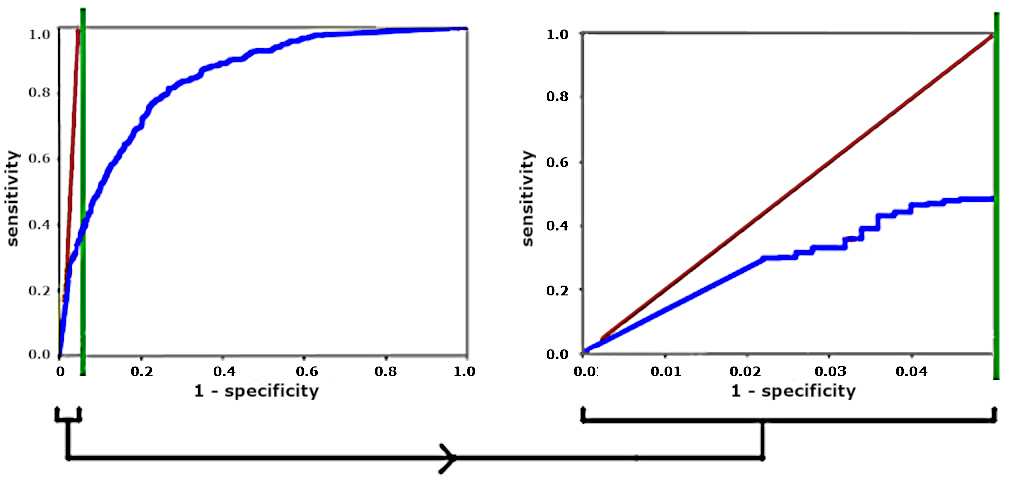}}
  \end{center}
{\caption{For a 20:1 distractor-to-parasite ratio, 
  stretching the left vertical sliver gives a more meaningful ROC curve.
  Diagonal red lines show operating points that give equal numbers of TPs and FPs.}
  \label{fig:misleadingRocs}
  }
\end{figure}

In such cases with large class imbalance (say $D$:1), the leftmost $\frac{1}{ D}$\textsuperscript{th} vertical sliver of the ROC curve, with \textit{y}-axis rescaled to be full width, reflects a more meaningful (and more sobering) ROC, because this expanded sliver visually weights TP counts and FP counts equally, as shown in Figure \ref{fig:misleadingRocs}. 

\begin{figure}[h!]
\begin{center}
{\includegraphics[width=0.8\textwidth]{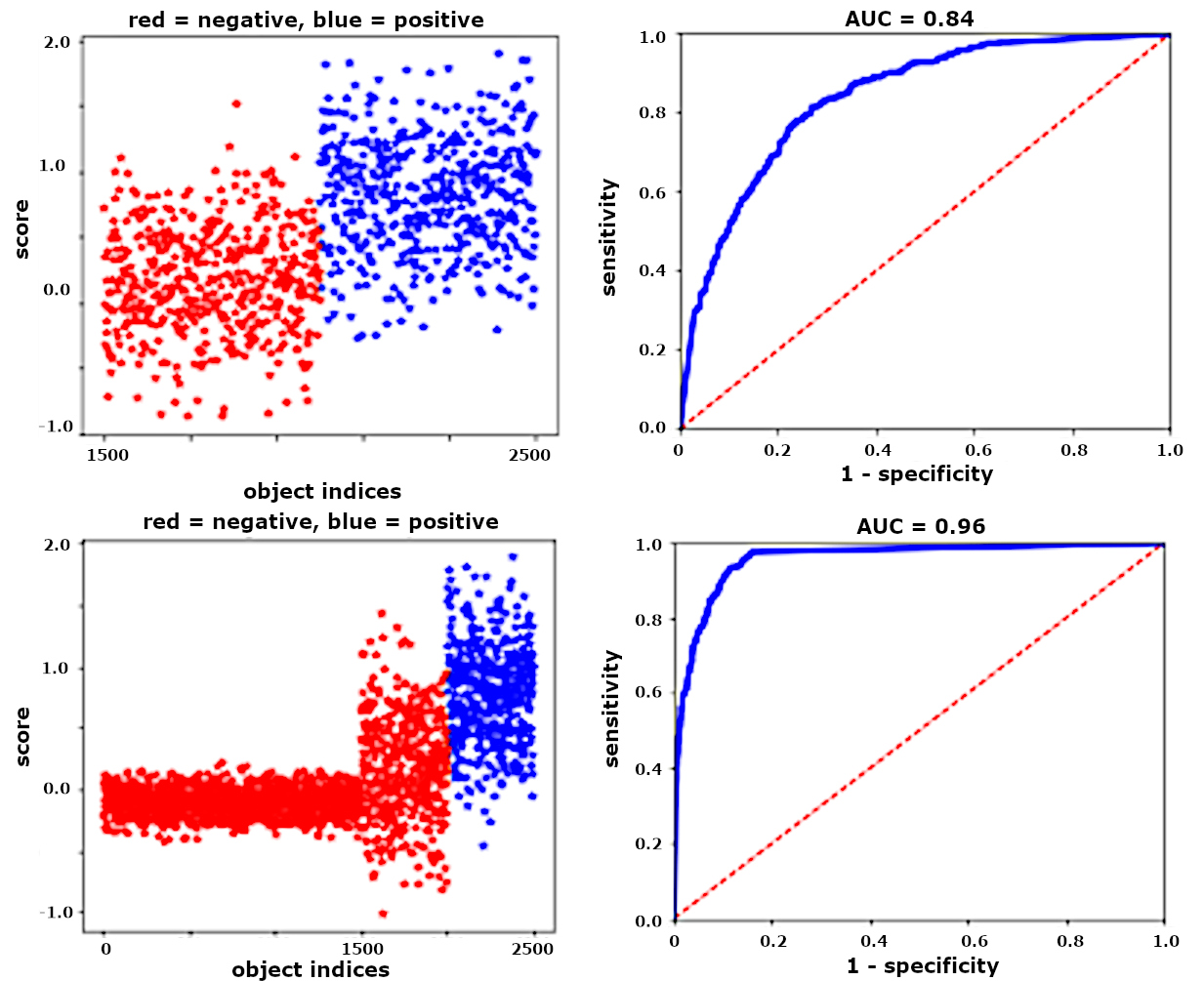}}
\end{center}
{\caption{For unchanged algorithm and FPR, ROCs are artificially improved by increasing the number of easy distractor objects.
Left: object scores (positives are blue, negatives are red).
Right: associated ROC curves.}
\label{fig:hardVsEasy}
}
\end{figure}

Third, the object-level ROC curve depends heavily on how distractors are defined.
For example, when using thick films to diagnose malaria, ``distractor" can mean (i) only the most difficult objects that closely resemble parasites; or (ii) any dark blob; or even (iii) every pixel in an image.  
Figure \ref{fig:hardVsEasy} shows an example in which considering only ``difficult" distractors (top) results in a low AUC, while considering  additional, mostly ``easy" distractors (bottom) gives a higher AUC with no change in actual performance as measured by FPR.
 
More informative than the object-level ROC is the Free ROC (FROC), which plots sensitivity vs FPR per $cV$.  
FROCs for object level are useful for development work: 
they clarify where gains can be made by favorably trading off object-level sensitivity for lower FP rates.
When datasets lack sufficient numbers of patients, FROCs on pooled objects can provide some insight into algorithm performance, with the caveat that they ignore patient-level variability.


\subsubsection{Patient-level ROCs}
\textit{Patient-level ROCs} can give a useful sense of algorithm behavior near the clinical performance requirements and are well worth reporting when sufficient data exists to plot them.  
However, there are two caveats. 
First, the only salient portion of a patient-level ROC is the region near clinically relevant operating points (e.g., specificity 90\%).
Second, because sensitivity is parasitemia-dependent (\ref{subsec:sensitivity}), the ROC is dependent also. 
Thus, a given algorithm may have much higher AUROC on a population with primarily high parasitemias than on one with lower parasitemias.

\subsubsection{Precision}
\textit{Object-level Precision} is the ratio of detected parasites over all detected objects, $\frac{tp}{tp + fp}$, and often appears as an ML metric.
This metric, as used, tends to badly underestimate the effects of parasite-to-distractor imbalances at the low LoDs required for clinical use, as follows. 

In ML papers, precision is often calculated on datasets with the clinically unrealistic situation of roughly balanced parasite and distractor counts, either because the numbers of objects have been artificially balanced or because the positive samples had high parasitemias (i.e. many parasites per volume $V$).
Since FPs roughly scale with volume $V$, high parasitemia samples yield much more balanced TP:FP ratios, which tend to give precisions which do not generalize to low parasitemia samples.

For example, a precision of 0.99 calculated on samples with $P \approx 10,000$ p/$\mu L$ corresponds to 100 FPs per $\mu L$ (assuming perfect sensitivity). 
At the required LoD of 100 p/$\mu L$, these same 100 FPs correspond to 100 parasites, giving precision = 0.5, a much less attractive result.

The related metric \textit{F1}, the harmonic mean of precision and object-level sensitivity (also problematic, as noted in \ref{subsec:sensitivity}), is a similarly misleading metric for reporting algorithm results, and in addition has no clinical utility. 

The rest of this section (\ref{subsec:patientLevelMetrics} - \ref{subsec:speciesIdMetrics}) discusses metrics that better reflect malaria's clinical use case.


\subsection{Patient level metrics}
\label{subsec:patientLevelMetrics}
The importance of assessing algorithm performance at the patient level cannot be over-emphasized. 
The basic unit of clinical care is the patient (we set aside population-level diagnostics such as for Vitamin A deficiency \cite{whoVitA}), so the most relevant metrics are defined at the patient level, not the object level.
Performance assessed across pooled objects can be a useful \textit{intermediate} step during ML development, but it is fundamentally unrealistic, because (i) it does not match the clinical task; (ii) it ignores interpatient variability; and (iii) it is dominated by high parasitemia samples. 
For example, consider four malaria-positive patients, with \\
$~~~~$Patient 1:           50,000 parasites/$\mu L$ (p/$\mu L$).   \\
$~~~~$Patients 2, 3, 4:  each 300 p/$\mu L$  \\
Suppose the algorithm detects all parasites in \{1\}, and misses all parasites in \{2,3,4\} (a realistic scenario due to interslide variability). 
Then the object-level sensitivity is 98\%, while patient-level sensitivity is 25\%.

We have found that two metrics, each defined on a per-patient basis, are particularly useful: \textit{false positive rate} (FPR) and \textit{sensitivity}.
Each is calculated separately for each patient, using algorithm accuracy on objects within that patient's sample.
These are covered in \ref{subsec:fpRate} and \ref{subsec:sensitivity}, and underpin other metrics related to specificity \ref{subsec:specificity}, LoD \ref{subsec:lod}, and quantitation \ref{subsec:quantitation}. 

\begin{figure}[h!]
\begin{center}
{\includegraphics[width=1\textwidth]{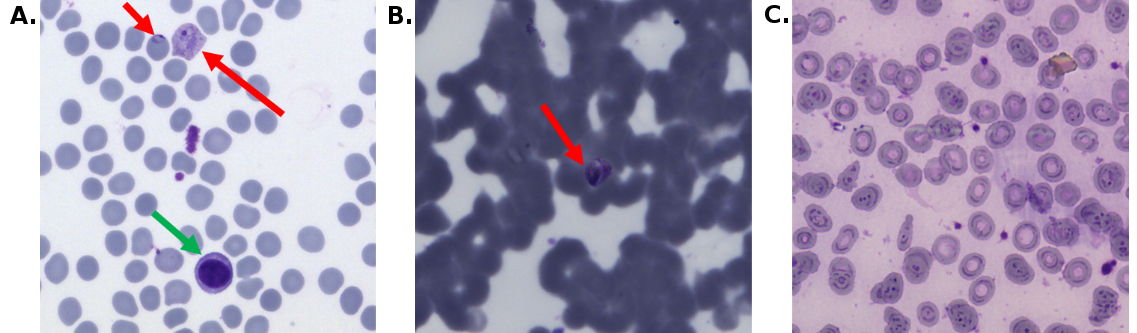}}
\end{center}
{\caption{Interpatient variability, thin blood films. Red arrows point to parasites. (c) is malaria-negative. By permission from \citep{thinGhtc2019}
}
\label{fig:intersampleVariability}
}
\end{figure}

Interpatient variability (as in Figure \ref{fig:intersampleVariability}) poses great difficulty for ML, so it must be factored into algorithm evaluation.
It is captured by the standard deviations of FPR and sensitivity (cf. \ref{subsec:fpRate}, \ref{subsec:sensitivity}), to the degree that the dataset captures interpatient diversity.

A related issue is interclinic variability.
For example, clinics can use different stain variants (e.g., Giemsa, Field, and JSB).
Even clinics with nominally identical protocols can differ substantially (see e.g., \cite{das} and a detailed example in \cite{torres}). 
Besides variations in presentation, different clinics may produce populations of samples with differently distributed FPRs and sensitivities.
Implications of this for tuning algorithms are covered in \ref{subsec:specificity}.
 
\subsection{FP Rate}
\label{subsec:fpRate}

False positive rate (FPR) is the number of distractors mislabeled as parasites per clinically relevant unit of substrate, hereafter $cV$, e.g., 1 $\mu L$ of blood (malaria), 10 mL urine (\textit{Schistosoma haematobium}), 1 gram stool (other \textit{Schisto}), a specified number of cells in a histological sample, etc; but \textit{not} ``per image tile", which generally has no clinical relevance (though image tiles can often be translated into the microscopy ``Fields of View" used in protocols). 
Malaria ML papers with some FPR analysis include \cite{linder, manescu2020, mehanian, thinGhtc2019}.
Crucially, FPR is calculated separately for each patient.
We denote the vector of FPRs for the population of patients as $\textbf{\textit{F}}$. 

FPR is \textit{not} object-level specificity, which is a commonly reported but highly flawed measure in this context (see \ref{subsec:problemsWithRocsAndAucs}).

While FPR can be calculated for any sample, FPRs on positive samples may be erroneously boosted by mis- or unannotated parasites. 
Thus,  the population's FPR distribution is best characterized using negative samples only.
 
Interpatient variability makes the standard deviation of FPR, $\sigma(\textbf{\textit{F}})$, a crucial performance measure. 
The mean FPR $\mu(\textbf{\textit{F}})$ is less relevant because it can be subtracted out, as shown in \ref{subsec:lod} and \ref{subsec:quantitation}. However, since it tends to scale roughly with $\sigma(\textbf{\textit{F}})$, it can give a hint as to the relative magnitude of $\sigma(\textbf{\textit{F}})$ (see e.g., \cite{mehanian, thinGhtc2019}).

In datasets with insufficient numbers of patients, an FPR calculated over pooled objects has some value as a lower bound on $\textbf{\textit{F}}$.
In particular, it can be compared to the clinical LoD requirement. 
For example, a pooled-object FPR of 5,000/$\mu L$, vs. a required 100 p/$\mu L$ LoD (malaria), is a clear sign that work is still needed.
Multiple splits of a set of pooled objects does not simulate $\sigma(\textbf{\textit{F}})$, because each split will include the full patient diversity.

\textit{Aside}: Samples with high FPRs are sometimes criticized as being due to ``poor sample preparation". 
However, except for extreme cases this is in the eye of the beholder: human clinicians readily and successfully diagnose ``dirty" samples on which ML algorithms fail. 
Thus, the need to improve sample prep is to large degree a need to accommodate ML methods' struggles with handling highly variable sample presentations.
See \cite{das}, and a detailed example in \cite{torres}. 


\subsection{Sensitivity}
\label{subsec:sensitivity}

Sensitivity (aka recall) is the fraction of positive items in a set that are correctly labeled:
Sensitivity = $\frac{tp}{tp + fn}$, $~$ where \textit{tp} = true positives, i.e. positive items labeled correctly, and \textit{fn} = false negatives, i.e. positive items labeled as negative or missed.
The ``items" can be parasites (object-level) or malaria-positive patients (patient-level).

\subsubsection{Pooled object sensitivity}

Sensitivity over a pooled set of parasites from multiple patients has some value as an \textit{intermediate} assessment metric during ML development (e.g., as a loss function for gradient descent training), if it is analysed carefully to avoid problems such as imbalanced parasitemias distorting the object pool (cf. the example given in \ref{subsec:patientLevelMetrics}).

\subsubsection{Per-patient object sensitivity}

A clinically realistic and useful version of object-level sensitivity measures each patient separately:\\ \\
\textit{Per patient object-level sensitivity} is the fraction of parasites in  the examined volume $V$ of a positive sample that are correctly labeled (e.g., by means of an object score threshold $C$), 
$\frac{tp}{tp + fn}$ $~$ where $tp$ = parasites labeled correctly, and $fn$ = parasites labeled as distractors (or missed). 
There is no constraint on the size of $V$ or parasitemia, but sensitivities for patients with few parasites are less reliable (cf. the law of large numbers).
Each patient's object-level sensitivity is calculated separately.
We denote the vector of sensitivities for the (malaria-positive) population as $\textbf{\textit{S}}$.
$\textbf{\textit{S}}$ underpins metrics related to LoD \ref{subsec:lod} and quantitation \ref{subsec:quantitation}.
 

\subsubsection{Patient-level sensitivity}

\textit{Patient-level sensitivity} is sensitivity in the usual clinical sense of the fraction of positive patients correctly diagnosed (not $\textbf{\textit{S}}$).
It is of course a vital metric clinically, but is complex to interpret because it depends on two things:

(i) The particular parasitemia distribution of the tested set: Patients with low  parasitemias (close to the LoD) are harder to identify. 
In malaria for example (where LoD $\approx$ 100 p/$\mu L$), if all patients have parasitemias $>$ 1000 p/$\mu L$, 100\% sensitivity is (hopefully) trivial, while if all parasitemias are under 50 p/$\mu L$, very low sensitivity is likely. 

(ii) The particular specificity: Sensitivity and specificity are paired and move in opposite directions, as seen in ROC curves. 

Thus, reporting patient-level sensitivity is uninformative and even misleading unless one also reports (i) the parasitemia distribution, and (ii) the associated specificity on negative samples.
The WHO competency levels are an important example:
These levels crucially assume the parasitemia distribution of the WHO 56 diagnosis slide set, viz 20 negative slides and 20 positive slides with parasitemias between 80 and 200 p/$\mu L$ \citep{whoQualityAssuranceV2}.
WHO competency level ratings do not apply to results on distributions with higher parasitemia samples.

A principled way to maximize patient-level sensitivity is given in \ref{subsec:optimizeSensitivity}.

\subsubsection{Effect of species on sensitivity}

Algorithm sensitivity results should be broken down by species as well as by parasitemia, because malaria species has strong impact on patient-level sensitivity.
This is due to the unique synchronization and sequestration behaviors of \textit{P. falciparum} \citep{garnham}: 

(i) In \textit{falciparum} the large, distinctive late stage forms sequester out of the peripheral blood, leaving only the smaller ring forms that are harder to detect and disambiguate from distractor objects (especially in thick films). 
As a result, non-\textit{falciparum} species (i.e. \textit{vivax, ovale, malariae, knowlesi}) are much easier to diagnose (given equal parasitemias), which allows an algorithm to have lower LoD and higher patient-level sensitivity. 

(ii) \textit{falciparum} parasites tend to synchronize in peripheral blood, with the presenting parasites forming a narrow age distribution. 
This strongly impacts diagnostic methods that target the biomarker hemozoin: non-\textit{falciparum} samples can be very sensitively detected due to the reliable presence of late-stage, high-hemozoin parasites \citep{arndt}, but even high parasitemia \textit{falciparum} samples can lack detectable hemozoin due to synchronized populations of early stage ring forms \citep{jamjoom, rebelo, delahuntHemozoin}, resulting in drastically different sensitivities by species. (Hemozoin appears to be a sensitive biomarker for \textit{falciparum} in cultured blood because synchronization is absent.)

This is a high-stakes issue because \textit{falciparum} is much more often fatal than non-\textit{falciparum} species.


\subsection{Specificity}
\label{subsec:specificity}
Specificity is the fraction of negative items (distractor objects or patients) that are correctly diagnosed as negative:  

Specificity = $\frac{tn}{tn + fp}$, $~$where $tn$ = true negatives (negative items in $V$ labeled correctly), and $fp$ = false positives (negative items labeled incorrectly).

 
\subsubsection{Object-level specificity} 

\textit{Object-level specificity}, even if calculated for each patient separately, has little usefulness and can be highly deceptive (see \ref{subsec:problemsWithRocsAndAucs}).

\subsubsection{Patient-level specificity}

\textit{Patient-level specificity}, i.e. in the usual clinical sense, is highly salient.
Clinical goals of high specificity include not overwhelming the health care system, avoiding excess treatments, and preventing misattribution.
Thus, clinical use-cases generally require a high specificity (eg 90\% for malaria diagnosis \cite{whoQualityAssuranceV2}, 97.5\% for schistosomiasis \citep{whoHelminths}).

Specificity is closely tied to FPR (\ref{subsec:fpRate} above) and can be readily tuned for an algorithm that labels objects:
Suppose that objects have been detected then labeled by some method (e.g., a threshold $C$ on object scores), that $\textbf{\textit{F}}$ (from \ref{subsec:fpRate}) is gaussian, and that patient diagnosis is determined by a threshold $T$ on the number of positively-labeled objects per $cV$  (i.e. a standard ``detect, classify, count, then threshold" approach). 
To attain a target specificity $K$, one can set 
\begin{equation}
T = \mu(\textbf{\textit{F}}) + \alpha~ \sigma(\textbf{\textit{F}}) 
\label{arbitrarySpecThresholdEqn}
\end{equation}
where  $\alpha$ is found via the (one-sided) error function and $K$. 
Alternate formulations for the case of non-gaussian $\textbf{\textit{F}}$ are given in section \ref{subsec:modifiedLod}.

Negative samples are easier to obtain and trivial to annotate (assuming accurate patient-level ground truth), and specificity depends only on negative samples.
So $T$ can ideally be tuned on a separate, dedicated validation set of negatives that capture a sufficient range of FPRs (both ``dirty" and ``clean" samples). 

Note that different clinics can have widely different FPR distributions $\textbf{\textit{F}}$.
Because $\sigma(\textbf{\textit{F}})$ determines both specificity (Eqn \ref{arbitrarySpecThresholdEqn}) and LoD \ref{subsec:lod}, different clinics may require different hyperparameters to hit the target patient specificity $K$, leading to different LoDs.
Thus, tuning an algorithm for deployment may involve multiple validation sets of negatives (by clinic), with clinic-dependent trade offs between specificity and higher LoD.


\subsection{Limit of detection (LoD)}
\label{subsec:lod}
Here, LoD roughly means the parasitemia at which the algorithm can consistently (e.g., 95\% of cases) distinguish positive and negative cases.
Based on the WHO evaluation criteria \cite{whoQualityAssuranceV2}, the required LoD for malaria microscopy is roughly 100 p/$\mu L$ , i.e. 1 parasite per 50,000 red blood cells (RBCs) or 80 white blood cells (WBCs). However, expert microscopists routinely achieve LoDs $\approx$50 p/$\mu L$ (e.g., \citep{DioniciaComment2015, davidBellComment}), and the lower LoD is of course clinically desirable. 
For helminths, LoD is implicitly 1 egg (per 10 mL urine or 1 gram stool) \cite{whoHelminths}.
 
LoD can be directly probed using holdout sets of low parasitemia positive samples.
These are not as useful for training anyway, as they supply few parasite objects. 
However, this is impractical because it's hard to acquire enough malaria samples near the LoD.

We can calculate a useful estimate of LoD from $\textbf{\textit{F}}$ and $\textbf{\textit{S}}$ as follows:

\textbullet $~$Denote the putative LoD as $L$ parasites per $cV$, and suppose that a patient is diagnosed as ``positive" when $N \ge T$, where $N$ is the number of positively-labeled objects per $cV$.  
Note that $N = TP + FP$ in positive patients, and $N = FP$ in negative patients, where $TP$ and $FP$ denote counts per $cV$, so $TP = tp ~ \frac{cV}{V}$ where $tp$ is the number of parasites correctly labeled in $V$ (similarly $FP = fp ~ \frac{cV}{V}$).
 
\textbullet $~$Make $T$ high enough to ensure to enforce 95\% specificity on negative samples as described in \citep{mehanian} by setting $\alpha$ to 1.65 std devs in Eqn \ref{arbitrarySpecThresholdEqn}:
\begin{equation}
T = \mu(\textbf{\textit{F}}) + 1.65 \sigma(\textbf{\textit{F}}) 
\label{thresholdEqn}
\end{equation} 

\textbullet $~$Then for positive samples the worst case is a very ``clean" sample with low FPR, such as the 5\textsuperscript{th} percentile of samples with $FP = \mu(\textbf{\textit{F}}) - 1.65 \sigma(\textbf{\textit{F}})$.
In this case we must depend mostly on detected parasites to ensure $N \geq T$ for a positive diagnosis.
Suppose for ease that the sample has average sensitivity = $\mu(\textbf{\textit{S}})$.
Then a sample at LoD has $TP = L\mu(\textbf{\textit{S}})$.

\textbullet $~$To diagnose this positive sample correctly (but just barely, i.e. $N=T$), we need
\begin{align*}
N = TP + FP &=  ~L\mu(\textbf{\textit{S}}) + \mu(\textbf{\textit{F}}) - 1.65 \sigma(\textbf{\textit{F}}) \\
&= ~ T = \mu(\textbf{\textit{F}}) + 1.65 \sigma(\textbf{\textit{F}})\\
&\Rightarrow  L\mu(\textbf{\textit{S}}) = 3.3 \sigma(\textbf{\textit{F}})
\label{lodDerivationEqn}
\end{align*}
So the estimated LoD ($L$ per $cV$) has
\begin{equation}
L = \frac{3.3 ~ \sigma(\textbf{\textit{F}})}{\mu(\textbf{\textit{S}})} 
\label{lodEqn}
\end{equation}

\textbullet $~$Optionally, $+1$ can be added to the numerator (i.e. require $N=T + 1$) to prevent unpredictable behavior should both $\sigma(\textbf{\textit{F}})$ and $\mu(\textbf{\textit{S}})$ approach 0:
\begin{equation}
L = \frac{3.3 ~ \sigma(\textbf{\textit{F}}) + 1}{\mu(\textbf{\textit{S}})} 
\label{lodPlusOneEqn}
\end{equation}

We have found this estimate to be a good (slightly optimistic) proxy for actual LoD when assessing algorithms during development.
It has the practical advantage that low parasitemia samples are unnecessary, because the vector $\textbf{\textit{S}}$ can be  well characterized by high parasitemia samples.
It also allows useful comparison of algorithms, as it directly addresses a key clinical requirement and is anchored to the relevant unit $cV$.

A more nuanced (and pessimistic) proxy could account for $\sigma(\bf{S})$ by having a denominator = $\mu(\textbf{\textit{S}}) - \beta ~ \sigma(\textbf{\textit{S}})$ for some $\beta$.

\subsection{Choosing operating points}
\label{subsec:optimizeSensitivity}
Given a trained algorithm that uses the two hyperparameters $C$ and $T$, $\{C, T\}$ can be optimized in a principled way to maximize patient-level sensitivity, subject to the constraint of a fixed target specificity $K$:  

\textbullet $~$Set aside a validation set of negative samples.
If there are sufficient positive samples to spare, optionally set these aside also.

\textbullet $~$For each $C$:\\ 
$~~~~~~~$- Calculate $\textbf{\textit{F}}$ over the validation negatives, and $\mu(\textbf{\textit{S}})$ over the validation positives if available, or (less ideal but workable) over the training set positives.\\
$~~~~~~~$- Determine $T = T(C, K, \textbf{\textit{F}})$ which hits the target specificity $K$ on the validation negatives, as in \ref{subsec:specificity}.
$~~~~~~~$- Estimate LoD as in \ref{subsec:lod}.

\textbullet $~$Select the $C$ with the lowest LoD.

\textbullet $~$Use this $\{C, T\}$ pair as algorithm hyperparameters to process test sets, and report
patient-level specificity and sensitivity. 

\subsection{Modified LoD and operating point formulas} 
\label{subsec:modifiedLod}

The methods for setting $T$ in \ref{subsec:specificity} and for estimating LoD  in \ref{subsec:lod} both assume that the FPR vector $\textbf{\textit{F}}$ is gaussian.
In our experience this is often not the case.
Rather, the FPR distribution may be asymmetrical, with mostly low-FPR samples and a few high-FPR samples.
This can be handled by modifying the methods in \ref{subsec:specificity} and \ref{subsec:lod} as follows:

\textbullet $~$For $\mu(\textbf{\textit{F}})$, use the median of $\textbf{\textit{F}}$  instead of the mean of $\textbf{\textit{F}}$.
    Similarly, if the vector $\textbf{\textit{S}}$ is non-gaussian, the median can be used instead of the mean for $\mu(\textbf{\textit{S}}$.

\textbullet $~$For $\sigma(\textbf{\textit{F}})$, use one-sided std devs, which can be calculated by keeping only the points to the right (or left) of the median and reflecting them across the median as centerpoint to create a symmetric distribution. 
This gives, for the FPR distribution above, a large right std dev $\sigma_R(\textbf{\textit{F}})$ and a small left std dev $\sigma_L(\textbf{\textit{F}})$.

\textbullet $~$Then the new versions of Eqns \ref{arbitrarySpecThresholdEqn} and \ref{lodEqn} are  
\begin{equation}
T = median(\textbf{\textit{F}}) + \alpha \sigma_R(\textbf{\textit{F}}) 
\label{threshold2Eqn}
\end{equation}
 
\begin{equation}
L = \frac{1.65 ~ (\sigma_L(\textbf{\textit{F}}) + \sigma_R(\textbf{\textit{F}}))}{\mu(\textbf{\textit{S}})}
\label{lod2Eqn}
\end{equation}
Two other methods of calculating $T$ from $\textbf{\textit{F}}$ may be useful:
\begin{enumerate}
    \item Set $T$ based on the $K$\textsuperscript{th} percentile of $\textbf{\textit{F}}$.
    \item Manually choose $T$ based on a scatterplot of the $FP$ counts in the validation negative samples.  
\end{enumerate}
For both these methods, the detected objects are assumed to be already classified.
If a threshold $C$ on object scores was used, then first $T$ must be calculated for each $C$, before choosing the best $\{C,T\}$ pair as in \ref{subsec:optimizeSensitivity}.

The manual method of choosing $\{C, T\}$ takes time, but it can yield the best results in a field deployment because it is most closely tailored to the empirical FPR distribution.
 

\subsection{Quantitation}
\label{subsec:quantitation}
Quantitation sometimes has clinical importance. 
For example, accurate quantitation is needed to monitor for drug-resistant malaria strains by calculating clearance curves \cite{ashley,white, whoMicroscopyQuantSOP}.
For helminths, quantitation targets are typically rough only (e.g., low, medium, high) \cite{whoHelminths}.
For \textit{Loa loa}, a remarkable drug reaction necessitates accurate quantitation at certain high parasitemias only ($\approx$ 20k to 30k worms/$\mu L$) \citep{gardon, dambrosio}.
\subsubsection{Measuring quantitation accuracy}
Quantitation accuracy should be reported at the patient level due to high interpatient variability.
For plotting quantitation error per patient, Bland-Altman plots are preferable because relative quantitation error is generally most important \citep{whoMalariaResearch2016}.

Reporting the $R^{2}$ value of a linear fit of estimated vs. true (i.e. $\hat{P}$ vs. $P$) is unsuitable when parasitemias range over orders of magnitude (common in malaria), because effects of the $L_2$ norm almost guarantee that high parasitemia samples will lay on the fitted line while high relative errors on low parasitemia samples will be downplayed, giving an illusion of strong fit.
Fitting the log($P$) rather than $P$ values helps to reduce this illusion. 
 \subsubsection{Estimating parasitemia}
As described in \citep{thinGhtc2019}, we can estimate the parasitemia $\hat{P}$ for a given patient by  
 \begin{equation}
 \hat{P}= \frac{n \left(\frac{cV}{V}\right) - \hat{\textbf{\textit{F}}}}{\hat{\textbf{\textit{S}}}}, \text{ where}
 \label{estQuantEqn}
 \end{equation}
$~~~~~~~$$n$ = number of alleged parasites found in $V$, \\ 
$~~~~~~~$$\hat{\textbf{\textit{F}}}$ = expected FPR (e.g., $~~~~~~~$$\mu(\textbf{\textit{F}})$),\\ 
$~~~~~~~$$\hat{\textbf{\textit{S}}}$ = expected sensitivity (e.g., $~~~~~~~$$\mu(\textbf{\textit{S}})$),\\
$~~~~~~~$$cV$ = clinically relevant volume of substrate, \\
$~~~~~~~$$V$ = estimate of the volume examined.

Three types of error affect Eqn \ref{estQuantEqn}: irreducible Poisson,  estimates of examined volume, and counts of  alleged parasites. \\ \\
\textit{Irreducible Poisson error:} \\
This is discussed below in \ref{subsec:effectOfPoissonStatistics}. \\ \\
\textit{Examined volume error:} \\
Error in estimating $V$ impacts quantitation accuracy via the $\frac{cV}{V}$ term of Eqn. \ref{estQuantEqn}. 
For example,  thick film blood volume $V$ is typically estimated by counting WBCs \citep{whoMicroscopy2016}.
Any error in the WBC count causes proportional quantitation error. 
This error type can be compartmentalized, for performance evaluation purposes only, as follows:

\textbullet $~$Manually count WBCs on a test set to ensure oracle $V$ estimates and use these counts to calculate $V$, ensuring zero error of this type.

\textbullet $~$Separately report the patient-level error statistics of the WBC counter. \\ \\
\textit{Parasite counting errors:} \\
Errors in parasite count stem from patient-level variations in sensitivity and FPR, as follows: 

\textbullet $~$The number of alleged parasites per $cV$ in the sample is $(tp + fp) \frac{cV}{V} = TP + FP$. 

\textbullet $~$Let $P$ is the true parasite count per $cV$.
Then $\hat{\textbf{\textit{S}}}P$ is the expected number of correctly labeled true parasites per $cV$, and the difference between $TP$ and $\hat{\textbf{\textit{S}}}P$ is due to deviation of the sample's sensitivity from the expected $\hat{\textbf{\textit{S}}}$.

\textbullet $~$Similarly, the difference between $FP$ and $\hat{\textbf{\textit{F}}}$ (the expected FPR) is due to the deviation of this sample's FPR from expected.
$\sigma(\textbf{\textit{S}})$ and $\sigma(\textbf{\textit{F}})$ quantify these deviations over the population.

\textbullet $~$A figure of merit to assess parasite counting error, derived and discussed in \citep{thinGhtc2019}, is thus
\begin{equation}
{\frac{\sigma(\textbf{\textit{S}})}{\mu (\textbf{\textit{S}})} } + {\frac{\sigma(\textbf{\textit{F}})}{\mu (\textbf{\textit{S}}) } \frac{1}{P} } 
\label{quantFomsEqn}
\end{equation}  
While the FPR term is usually hardest to control, it also shrinks as  $1/P$, so for large $P$ the sensitivity  term dominates.
This effect can be leveraged by using different operating points according to whether initial estimated parasitemia is low or high, to favor FPR or sensitivity.
In particular, different operating points are indicated for diagnosis (since the hard cases have low parasitemia, where FPR dominates) and for quantitation (high parasitemias, where sensitivity dominates).

We note that parasitemia estimates based on manual microscopy are also subject to these three error types.
This complicates assessment of a model's quantitation accuracy against microscopy ground truth.
 
 
\subsection{Effect of Poisson statistics}
\label{subsec:effectOfPoissonStatistics}
Poisson statistics for rare events give variation in the actual number of parasites in a particular sample with volume $V$, given a fixed true parasitemia $P$ over the whole sample. 
The variation is most visible at low parasitemias, e.g., at 100 p/$\mu L$, where each RBC has a 1/50,000 chance of containing a parasite in thin film, or each WBC has a 1/80 chance of corresponding to a nearby parasite in thick film.  

\begin{figure}[h!]
\begin{center}
  {\includegraphics[width=0.8\textwidth]{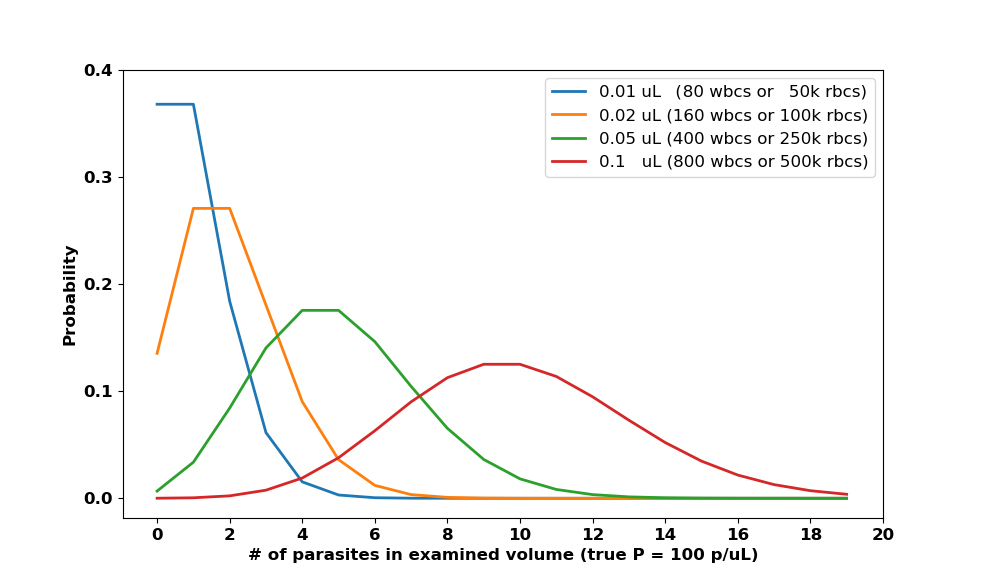}}
  \end{center}
  {\caption{Poisson distributions of parasite counts for various examined volumes $V$, assuming 100 p/$\mu L$. Low parasitemia samples may present as negative (i.e. zero parasites) if $V$ is too small.}
  \label{fig:poissonDist} 
}
\end{figure}

This variability has two main impacts:

(i) For diagnosis, a low LoD requires that a large volume $V$ be examined to ensure that at least a couple true parasites are present at all.
Otherwise, for a statistically predictable subset of positive patients the examined volume will contain 0 parasites, reducing patient-level sensitivity from the start.
For malaria, with LoD of 100 p/$\mu L$, this means that at least $\approx$0.05 $\mu L$ of blood should be examined, equivalent to 200 WBCs in thick film, or 250,000 RBCs in thin film (the difficulty of finding this many acceptable RBCs, and the long processing time required, are two reasons why thin films are not standard protocol for manual field diagnosis).
The miLab platform \citep{noul} examines $\approx$ 200k RBCs on thin film (close to the green curve in Fig \ref{fig:poissonDist}), and the Autoscope/EasyScan GO \citep{mehanian, das} examines $>$0.1 $\mu L$ of thick film (the red curve in Fig \ref{fig:poissonDist}).

(ii) For quantitation, a sufficiently high volume $V$ (depending on $P$) must be examined to control irreducible error.
For more detail see S.I. of \cite{thinGhtc2019}.

In both cases, automated systems hold a strong advantage because they can scan higher volumes than human technicians, who often by necessity work in a high Poisson error regime (S.I. of \cite{thinGhtc2019}).
Manual microscopy protocols average multiple readers' estimates (when available) to reduce quantitation error \citep{whoMalariaResearch2016, obare}.
 
When reporting results on datasets of small size, authors should understand how Poisson variability limits their estimates of algorithm performance. 

\subsection{Malaria species identification metrics}
\label{subsec:speciesIdMetrics}

Identification of malaria species is one of the three tasks assessed by the WHO 56 evaluation system \citep{whoQualityAssuranceV2}.
Correct species ID matters clinically because (i) \textit{falciparum} infections are much more likely to be fatal; and (ii) treatment plans differ by species \citep{cdcMalariaTreatment}, since (for example) the hypnozoites of \textit{vivax} and \textit{ovale} species require special care.

Because not all species ID errors are equal from a clinical perspective, reported results should preferably include a confusion matrix as in \citep{thinGhtc2019}.

Aside: It is relatively straightforward to distinguish \textit{falciparum} vs non-\textit{falciparum} on thick film alone \citep{vongpromek, torres, das, kassim}, and even mixed species infections that include \textit{falciparum} can often be identified on thick film by comparing the ring stage and late stage parasite counts \citep{horningWho55}.
However, thin films are still typically needed to distinguish between the various non-\textit{falciparum} species, unless the clinical use case allows geographical priors to be leveraged. 
A method to distinguish non-\textit{falciparum} species on thick film would yield clinical benefit by eliminating the need for thin films, due to (i) the ease of thick-only workflows \cite{janeCarterComment, stephaneComment}, and (ii) thin film problems with quality \citep{earlLongComment2015} and difficulty of species ID at  low parasitemias (\citep{KenPersonalCommunication2015}).
 
Staging parasites (as ring, later trophozoite, schizont, or gametocyte) is not part of the WHO evaluation, and is not generally useful clinically, except as used during species identification or when quantitating asexual forms in non-\textit{falciparum} species.
In \textit{falciparum} (the main target of quantitation, for drug resistance studies) the difference between ring and gametocyte is glaring.


\section{Discussion}
\label{sec:discussion}

Malaria and NTDs are amenable though difficult targets for ML methods, and successful development of translatable ML solutions would yield tremendous health care benefits for currently underserved populations.

Unfortunately communal ML progress, in which researchers build on each others' work to reach a performance goal, is handicapped for malaria by lack of attention to clinical needs, and by widespread use of ill-suited evaluation metrics.
As a result, the synergistic power of the ML community is not being applied with full force to this important task, as many papers present methods that cannot be usefully extended. 
 
Individual ML research teams can radically improve the situation by grounding their ML work in an understanding of the use case, and by tailoring metrics to the clinical needs.
We have described such metrics here: variation in FPR, per-patient sensitivity, LoD, patient-level sensitivity and specificity, and a figure of merit for quantitation.
We have also listed some essential technical background reading from the WHO and others.

Peer reviewers play a special role in determining the success or failure of the communal ML effort: 
(i) reviewers can assess algorithms and performance results according to whether they incorporate the requirements of the clinical use case.
(ii) when authors present new metrics, well-grounded in the use-case, this can be more valuable than a comparison based on customary but inferior metrics. 
By recognizing when this is the case, reviewers can disrupt the cycle that perpetuates a counterproductive status quo.

With attention to the clinical use case and deliberate choice of metrics, the ML community can better equip itself to successfully address automated malaria and NTD diagnosis, and thus deliver concrete benefit to the populations suffering the dire effects of these illnesses. 






\section*{Funding}
Funding provided by Global Health Labs, Inc. (\url{www.ghlabs.org})




\bibliographystyle{ieeeTran}
\bibliography{metricsBibliography}

\begin{thebibliography}{10}
\providecommand{\url}[1]{#1}
\csname url@samestyle\endcsname
\providecommand{\newblock}{\relax}
\providecommand{\bibinfo}[2]{#2}
\providecommand{\BIBentrySTDinterwordspacing}{\spaceskip=0pt\relax}
\providecommand{\BIBentryALTinterwordstretchfactor}{4}
\providecommand{\BIBentryALTinterwordspacing}{\spaceskip=\fontdimen2\font plus
\BIBentryALTinterwordstretchfactor\fontdimen3\font minus
  \fontdimen4\font\relax}
\providecommand{\BIBforeignlanguage}[2]{{%
\expandafter\ifx\csname l@#1\endcsname\relax
\typeout{** WARNING: IEEEtran.bst: No hyphenation pattern has been}%
\typeout{** loaded for the language `#1'. Using the pattern for}%
\typeout{** the default language instead.}%
\else
\language=\csname l@#1\endcsname
\fi
#2}}
\providecommand{\BIBdecl}{\relax}
\BIBdecl

\bibitem{whoMalaria2019}
{WHO}, \emph{World malaria report 2019}, 2019, {World Health Organization,
  Geneva, Switzerland}.

\bibitem{gfNTDs}
{BMGF}, ``{Neglected tropical diseases},'' {2023},
  \url{https://www.gatesfoundation.org/our-work/programs/global-health/neglected-tropical-diseases}.

\bibitem{metricsReloaded}
L.~Maier-Hein, A.~Reinke, and et~al., ``Metrics reloaded: Pitfalls and
  recommendations for image analysis validation,'' \emph{arXiv}, 2022,
  \url{https://arxiv.org/abs/2206.01653}.

\bibitem{draredech}
{draredech}, website, \url{https://draradech.github.io/jigsaw/jigsaw-hex.html}.

\bibitem{ghtc2014}
C.~Delahunt, S.~McGuire, M.~Horning, and et~alia, ``Automated microscopy and
  machine learning for expert-level malaria field diagnosis,'' \emph{IEEE GHTC
  Proceedings}, 2014.

\bibitem{mehanian}
C.~Mehanian, M.~{Horning}, and et~al., ``Computer-automated malaria diagnosis
  and quantitation using convolutional neural networks,'' \emph{ICCV}, 2017.

\bibitem{thinGhtc2019}
C.~Delahunt, M.~Jaiswal, C.~Mehanian, and et~al., ``Fully-automated
  patient-level malaria assessment on field-prepared thin blood film microscopy
  images,'' \emph{IEEE GHTC Proceedings}, 2019, (with S.I.)
  \url{https://arxiv.org/abs/1908.01901}.

\bibitem{torres}
K.~Torres, C.~Bachman, V.~Gamboa, and et~al., ``Automated microscopy for
  routine malaria diagnosis: a field comparison on {G}iemsa-stained blood films
  in {P}eru,'' \emph{Malaria J}, 2018.

\bibitem{vongpromek}
R.~Vongpromek, S.~Proux, L.~Ekawati, L.~Archasuksan, C.~Bachman, D.~Bell, and
  et~al., ``Field evaluation of automated digital malaria microscopy: {EasyScan
  GO},'' \emph{Trans R Soc Trop Med Hyg.}, 2019.

\bibitem{horningWho55}
M.~Horning, C.~Delahunt, C.~Bachman, and et~al., ``Performance of a
  fully‐automated system on a {WHO} malaria microscopy evaluation slide
  set,'' \emph{Malaria J}, 2021.

\bibitem{das}
D.~Das, R.~Vongpromed, M.~Dhorda, and et~al., ``Field evaluation of the
  diagnostic performance of {EasyScan GO}: a digital malaria microscopy device
  based on machine-learning,'' \emph{Malaria J}, 2022.

\bibitem{reesChanner}
R.~Rees-Channer, C.~Bachman, P.~Chiodini, and et~al, ``Evaluation of an
  automated microscope using machine learning for the detection of malaria in
  travelers returned to the {UK},'' \emph{In review}, 2023.

\bibitem{wiens}
J.~Weins, S.~Saria, A.~Goldenberg, and et~al., ``Do no harm: a roadmap for
  responsible machine learning for health care,'' \emph{Nature Medicine}, 2019.

\bibitem{koller}
D.~Koller and Y.~Bengio, ``A fireside chat with {D}aphne {K}oller at {ICLR},''
  2018, \url{https://www.youtube.com/watch?v=N4mdV1CIpvI}.

\bibitem{whoEvidenceForAi}
{WHO}, \emph{Generating evidence for artificial intelligence-based medical
  devices: a framework for training, validation and evaluation}, 2021, {World
  Health Organization, Geneva, Switzerland}.

\bibitem{whoMicroscopyLearners2010}
------, \emph{Basic malaria microscopy. Part I. Learner’s guide. 2nd ed, (esp
  units 7, 8, and 9)}, 2010, {World Health Organization, Geneva, Switzerland}.

\bibitem{whoMicroscopy2016}
------, \emph{Microscopy examination of thick and thin blood films for
  identification of malaria parasites (esp SOPs 8 and 9)}, 2016, {World Health
  Organization, Geneva, Switzerland}.

\bibitem{peruMalaria}
{Ministerio de Salud}, ``{Manual de Procedimientos de Laboratoria Para el
  Diagnostico de Malaria},'' 2003, {Lima, Peru}.

\bibitem{cdcMalaria}
{CDC}, ``Malaria website,'' {2023}, {Centers for Disease Control}
  \url{https://www.cdc.gov/malaria/index.html}.

\bibitem{whoMalariaResearch2016}
{WHO}, ``Microscopy for the detection, identification and quantification of
  malaria parasites on stained thick and thin blood films in research settings,
  ver 1,'' 2016, {World Health Organization, Geneva, Switzerland}.

\bibitem{whoQualityAssuranceV2}
------, ``Malaria microscopy quality assurance manual v2,'' 2016, {World Health
  Organization, Geneva, Switzerland}.

\bibitem{whoQualityAssuranceV1}
------, \emph{Malaria Microscopy Quality Assurance Manual V1}, 2009, {World
  Health Organization, Geneva, Switzerland}.

\bibitem{whoHelminths}
------, ``Diagnostic target product profiles for monitoring, evaluation and
  surveillance of schistosomiasis control programmes,'' 2021, {World Health
  Organization, Geneva, Switzerland}.

\bibitem{oyiboSchisto2022}
\BIBentryALTinterwordspacing
P.~Oyibo, S.~Jujjavarapu, J.~Diehl, and et~al., ``Schistoscope: An automated
  microscope with artificial intelligence for detection of
  \textit{{S}chistosoma haematobium} eggs in resource-limited settings,''
  \emph{Micromachines}, 2022. [Online]. Available:
  \url{https://www.mdpi.com/2072-666X/13/5/643}
\BIBentrySTDinterwordspacing

\bibitem{manescu2020}
P.~Manescu, M.~Shaw, D.~Fernandez-Reyes, and et~al., ``Expert-level automated
  malaria diagnosis on routine blood films with deep neural networks,''
  \emph{American Journal of Hematology}, 2020.

\bibitem{manescuWeaklySupervised}
P.~Manescu, C.~Bendkowski, D.~Fernandez-Reyes, and et~al, \emph{A Weakly
  Supervised Deep Learning Approach for Detecting Malaria and Sickle Cells in
  Blood Films}, 2020.

\bibitem{kassim}
Y.~Kassim, F.~Yang, H.~Yu, R.~Maude, and S.~Jaeger, ``Diagnosing malaria
  patients with \textit{{P}lasmodium falciparum} and \textit{vivax} using deep
  learning for thick smears,'' \emph{Diagnostics}, 2021.

\bibitem{poostchi2022}
M.~Poostchi, S.~Jaeger, and et~al, ``Malaria parasite detection and cell
  counting for human and mouse using thin blood smear microscopy,'' \emph{J
  Medical Imaging}, 2022.

\bibitem{yang2020}
F.~Yang, M.~Poostchi, H.~Yu, Z.~Zhou, K.~Silamut, J.~Yu, and et~al., ``Deep
  learning for smartphone-based malaria parasite detection in thick blood
  smears,'' \emph{IEEE J Biomed Health Inform}, 2020.

\bibitem{yu2023}
H.~Yu, O.~Fayad, S.~Jaeger, and et~al, ``Patient-level performance evaluation
  of a smartphone-based malaria diagnostic application,'' \emph{Malaria J},
  2023.

\bibitem{armstrong}
M.~Armstrong, J.~Coulibaly, S.~Essien-Baidoo, I.~Bogoch, D.~Fletcher, and
  et~alia, ``Point-of-care sample preparation and automated quantitative
  detection of \textit{{S}chistosoma haematobium} using mobile phone
  microscopy,'' \emph{Am J Trop Med Hyg}, 2022.

\bibitem{poostchi}
M.~Poostchi, K.~Silamut, R.~Maude, S.~Jaeger, and G.~Thoma, ``Image analysis
  and machine learning for detecting malaria,'' \emph{Translational Research},
  2018.

\bibitem{earlLongComment2015}
E.~Long, ``{(CDC) Personal communication},'' 2015.

\bibitem{noul}
Noul, ``{miLab} platform,'' {2023}, s. Korea.
  \url{https://noul.kr/en/milab-platform/}.

\bibitem{prakeshThinFilmSpreader}
\BIBentryALTinterwordspacing
J.~Nowak, A.~Kothari, H.~Li, J.~Pannu, D.~Algazi, and M.~Prakash, ``Inkwell:
  Design and validation of a low-cost open electricity-free 3d printed device
  for automated thin smearing of whole blood,'' \emph{arXiv}, 2023. [Online].
  Available: \url{https://arxiv.org/abs/2304.10200}
\BIBentrySTDinterwordspacing

\bibitem{whoVitA}
{WHO}, ``Serum retinol concentrations for determining the prevalence of vitamin
  a deficiency in populations,'' 2011, {World Health Organization, Geneva,
  Switzerland}.

\bibitem{linder}
N.~Linder and et~al., ``A malaria diagnostic tool based on computer vision
  screening and visualization of \textit{{P}lasmodium falciparum} candidate
  areas in digitized blood smears,'' \emph{PLoS One}, 2014.

\bibitem{garnham}
P.~Garnham, \emph{Malaria parasites and other haemosporidia}.\hskip 1em plus
  0.5em minus 0.4em\relax Blackwell Scientific Publications Ltd., 1966.

\bibitem{arndt}
L.~Arndt, T.~Koleala, A.~Orban, I.~Kezsmarki, S.~Karl, and et~alia,
  ``Magneto-optical diagnosis of symptomatic malaria in papua new guinea,''
  \emph{Nature Communications}, 2021.

\bibitem{jamjoom}
G.~Jamjoom, ``Patterns of pigment accumulation in \textit{{P}lasmodium
  falciparum} trophozoites in peripheral blood samples,'' \emph{Am J Trop Med
  Hyg}, 1988.

\bibitem{rebelo}
M.~Rebelo, H.~Shapiro, T.~Amaral, J.~Melo-Cristino, and T.~Hanscheid,
  ``Haemozoin detection in infected erythrocytes for \textit{{P}lasmodium
  falciparum} malaria diagnosis-prospects and limitations,'' \emph{Acta
  Tropica}, 2011.

\bibitem{delahuntHemozoin}
C.~Delahunt, M.~Horning, B.~Wilson, J.~Proctor, and M.~Hegg, ``Limitations of
  haemozoin-based diagnosis of \textit{{P}lasmodium falciparum} using
  dark-field microscopy,'' \emph{Malaria J}, 2014.

\bibitem{DioniciaComment2015}
D.~G. Vilela, ``{(Universidad Peruana Cayetano Heredia) Personal
  communication},'' 2015.

\bibitem{davidBellComment}
D.~Bell, ``{(WHO, FIND, Global Good) Personal communication},'' 2016.

\bibitem{ashley}
E.~Ashley, M.~Dhorda, R.~Fairhurst, C.~Amaratunga, and et~al., ``Spread of
  artemisinin resistance in \textit{{P}lasmodium falciparum} malaria,''
  \emph{New England J of Medicine}, 2014.

\bibitem{white}
N.~White, ``The parasite clearance curve,'' \emph{Malaria J}, 2011.

\bibitem{whoMicroscopyQuantSOP}
{WHO}, \emph{Malaria Microscopy Standard Operating Procedure MM-SOP-09: Malaria
  Parasite Counting}, 2016, {World Health Organization, Geneva, Switzerland}.

\bibitem{gardon}
J.~Gardon, M.~Boussinesq, and et~al., ``Serious reactions after mass treatment
  of onchocerciasis with ivermectin in an area endemic for loa loa infection,''
  \emph{Lancet}, 1997.

\bibitem{dambrosio}
M.~V. D’Ambrosio, D.~Fletcher, and et~al., ``Point-of-care quantification of
  blood-borne filarial parasites with a mobile phone microscope,''
  \emph{Science Trans Medicine}, 2015.

\bibitem{obare}
{WWARN}, ``Obare method calculator,'' {2023},
  \url{https://www.wwarn.org/obare-method-calculator}.

\bibitem{cdcMalariaTreatment}
{CDC}, ``Algorithm for diagnosis and treatment of malaria in the {U}nited
  {S}tates,'' {2023}, {Centers for Disease Control}
  \url{https://www.cdc.gov/malaria/resources/pdf/Malaria_Managment_Algorithm_202208.pdf}.

\bibitem{janeCarterComment}
J.~Carter, ``{(Amref) Personal communication},'' 2015.

\bibitem{stephaneComment}
S.~Proux, ``{(SMRU) Personal communication},'' 2015.

\bibitem{KenPersonalCommunication2015}
K.~Lilley, ``{(Australian Defence Force Malaria and Infectious Disease
  Institute) Personal communication},'' 2015.

\end{thebibliography}

\end{document}